\documentclass[runningheads]{llncs}

\usepackage{eccv}

\usepackage{eccvabbrv}

\usepackage{graphicx}
\usepackage{booktabs}

\usepackage[accsupp]{axessibility}  % Improves PDF readability for those with disabilities.

\usepackage{amssymb}
\usepackage{hyperref}

\usepackage{orcidlink}
\usepackage{arydshln} 
\usepackage{multirow}
\usepackage{colortbl}
\usepackage{pifont}
\usepackage{colortbl}
\usepackage{xcolor}
\usepackage{wrapfig}

\begin{document}

\title{Multi-Grained Cross-modal Alignment for Learning Open-vocabulary Semantic Segmentation from Text Supervision} 

\titlerunning{Multi-Grained Cross-modal Alignment}

\author{Yajie Liu\inst{1} \and
Pu Ge\inst{2} \and
Qingjie Liu\inst{1} \and
Di Huang \inst{1}}

\authorrunning{Y.~Liu et al.}

\institute{Beihang University, Beijing, China, \and
Hangzhou Innovation Institute, Hangzhou, China}

\maketitle

\begin{abstract}
  % The abstract should summarize the contents of the paper. 
  % LNCS guidelines indicate it should be at least 70 and at most 150 words.
  % Please include keywords as in the example below. 
  % This is required for papers in LNCS proceedings.
 
Recently, learning open-vocabulary semantic segmentation from text supervision has achieved promising downstream performance.
Nevertheless, current approaches encounter an alignment granularity gap owing to the absence of dense annotations, wherein they learn coarse image/region-text alignment during training yet perform group/pixel-level predictions at inference. 
% that necessitate finer alignment. 
Such discrepancy leads to suboptimal learning efficiency and inferior zero-shot segmentation results.
In this paper, we introduce a Multi-Grained Cross-modal Alignment (MGCA) framework, which explicitly learns pixel-level alignment along with object- and region-level alignment to bridge the granularity gap without any dense annotations.
Specifically, MGCA ingeniously constructs pseudo multi-granular semantic correspondences upon image-text pairs and collaborates with hard sampling strategies to facilitate fine-grained cross-modal contrastive learning.
Further, we point out the defects of existing group and pixel prediction units in downstream segmentation and develop an adaptive semantic unit which effectively mitigates their dilemmas including under- and over-segmentation. 
Training solely on CC3M, our method achieves significant advancements over state-of-the-art methods, demonstrating its effectiveness and efficiency. 

\end{abstract}

\section{Introduction}
\label{sec:intro}

Open-vocabulary semantic segmentation (OVSS) aims to segment the visual elements of arbitrary categories. In contrast to conventional semantic segmentation~\cite{zhou2019semantic,chen2018encoder} that segments targets in a fixed set of categories, OVSS is more practical for real-world applications where infinite visual concepts exist.  
However, the expensive pixel-level annotations and restricted label categories impede the learning of arbitrary categories.
Motivated by the remarkable success of visual-language pre-training (VLP)~\cite{radford2021learning,jia2021scaling}, recent studies~\cite{xu2022groupvit,luo2023segclip,cha2023learning,xing2023rewrite,ren2022viewco,mukhoti2023open,liu2022open, xu2023learning} propose to learn OVSS from massive web-crawled image-text paired data, which obviates the need for expensive manual pixel-level annotations. 
Despite the impressive progress, there remain critical unresolved challenges that hinder the efficient learning of dense cross-modal alignment.  

\begin{figure*}[tb]
\centering
\includegraphics[scale=0.49]{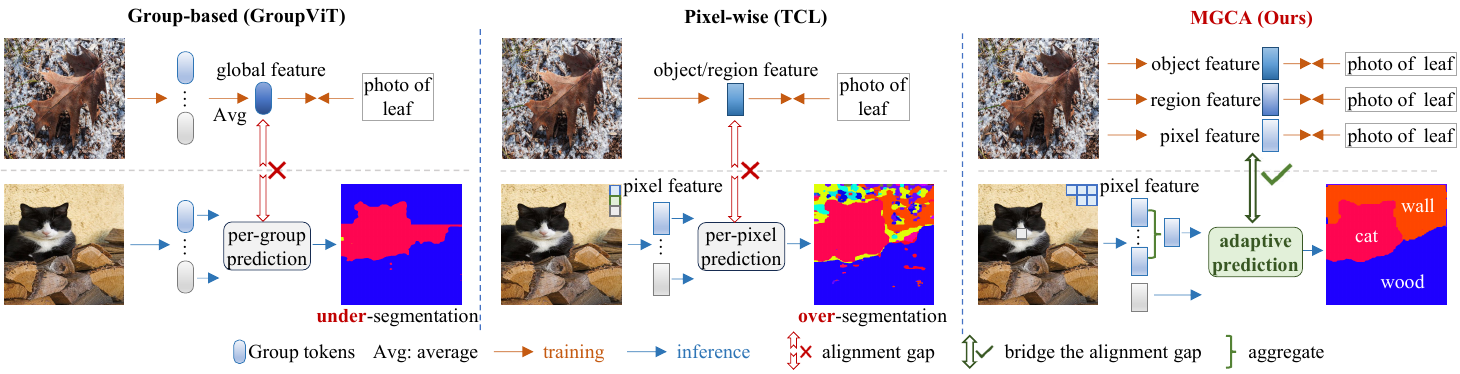}
\vspace{-15pt}
\caption{Conceptual comparison between previous methods and ours. Group-based methods encounter an image-group alignment gap while pixel-wise methods confront a region-pixel alignment gap. In contrast, our method bridges the train-test alignment gap and introduces adaptive prediction that mitigates under-/over-segmentation issues. }    
\vspace{-15pt}
\label{fig:problem}
\end{figure*}

The primary challenge lies in the alignment granularity gap, as the task learns from coarse image-level supervision yet necessitates precise predictions for dense pixels.
To learn segmentation with text supervision, the mainstream works adopt two typical pipelines.
The first is group-based methods~\cite{xu2022groupvit,ren2022viewco,luo2023segclip}, which explicitly cluster an image into several group tokens. Groups are averaged to align with image-level supervision, but at inference, segmentation masks are generated by independently matching each group with texts.
The \textcolor{blue}{image}-\textcolor{magenta}{group} alignment gap between \textcolor{blue}{training} and \textcolor{magenta}{inference} (see \cref{fig:problem}) leads to limited performance in complex scenes despite extensive training on large-scale data (\eg, 8.7 mIoU on ADE20K~\cite{zhou2019semantic} with 29 million image-text pairs in ~\cite{xu2022groupvit}).
The other we refer to pixel-wise methods~\cite{cha2023learning,mukhoti2023open} that strive to achieve fine-grained alignment without the need for pixel reorganization.
More specifically, TCL~\cite{cha2023learning} incorporates a text grounder into CLIP~\cite{radford2021learning} to learn region-level alignment, yet performs pixel-wise classification at inference. 
PACL~\cite{mukhoti2023open} establishes alignment between texts and attention weighted global visual embeddings but conducts pixel-level predictions.
Even with the rich knowledge of CLIP, pixel-wise methods require heavily fine-tuning on massive paired data (15 million in ~\cite{cha2023learning}) to achieve promising results due to the \textcolor{blue}{train}-\textcolor{magenta}{test} \textcolor{blue}{region}-\textcolor{magenta}{pixel} alignment gap as depicted in \cref{fig:problem}.

Another critical issue is rarely explored regarding the semantic unit in downstream segmentation inference. Semantic units refer to the basic visual elements processed by the recognition module, which means that pixels within a unit share the same category. However, our observation reveals that existing units impose limitations on segmentation performance. Group-based methods recognize each group unit, which is an object-level representation formed with limited global initial centroids and without explicit group-level supervision. As a result, the group unit suffers from imprecise grouping and under-segmentation at zero-shot inference as shown in \cref{fig:problem}. In contrast, pixel-wise methods classify each pixel independently and disregard the semantic-level visual coherence.
However, without sufficient pixel-level supervision, individual pixel embedding may not capture adequate semantics and be susceptible to noise, resulting in over-segmentation, \ie, inconsistent semantics appear within an object as illustrated in \cref{fig:problem}.

To bridge the alignment gap, we propose a Multi-Grained Cross-modal Alignment (MGCA) framework that innovatively constructs  multi-granular pseudo semantic correspondence for fine-grained alignment. 
As illustrated by the provided examples in the right of \cref{fig:overall}, the text captions demonstrate a stronger correspondence with the specific objects/regions/pixels in paired images compared to those in unpaired. Current methods are constrained to the object/region-level alignment, failing to capture the most critical pixel-level correspondence that aligns with the semantic granularity required for segmentation tasks. 
In contrast, MGCA establishes the informative object/region/pixel-level positive and negative pairs based on the pixel-to-text similarity matrix and subsequently conducts multi-grained cross-modal contrastive learning, explicitly resolving the alignment granularity gap and enhancing the recognition capabilities at various granularity.
Moreover, to tackle the redundant matching candidates present in region/pixel-level alignment, MGCA introduces efficient hard mining strategies which facilitate better and faster learning with zero computational overhead.

To alleviate the under- and over-segmentation problems at inference, we construct an adaptive semantic unit which cleverly harnesses the advantages of existing group and pixel unit.
The core idea is to finely aggregate the semantically relevant pixels centered on adequate local sampling points to form valid part-level representations. The aggregation dynamically exploits the generalizable visual coherence~\cite{mukhoti2023open} to enforce more consistent predictions and effectively suppress the adverse impact of inconsistent noise. Furthermore, our semantic unit adaptively harnesses the learned multi-granular alignment capabilities for dense predictions. Our main contributions can be summarized as follows:
\begin{itemize}
    \item We propose an efficient Multi-Grained Cross-modal Alignment (MGCA) framework for text-supervised open-vocabulary semantic segmentation. It bridges the train-test alignment gap and enables adaptive predictions. 
    \item We innovatively establish pseudo semantic correspondence at object-, region- and pixel-level for fine-grained alignment without any dense annotations.
    \item We develop an adaptive semantic unit for downstream segmentation, which mitigates the under-/over-segmentation issues arising from group/pixel units.
    \item Training solely on CC3M datasets~\cite{sharma2018conceptual} with mere 4.72M learnable parameters, we achieve new state-of-the-art zero-shot performance on 8 segmentation benchmarks with large margins compared to previous methods. 
\end{itemize}

\section{Related Work}
\label{sect:related work}

\textbf{Open-vocabulary semantic segmentation with text supervision.} Motivated by the remarkable success of VLP~\cite{radford2021learning,jia2021scaling}, recent studies~\cite{xu2022groupvit,luo2023segclip,cha2023learning,xing2023rewrite,ren2022viewco,mukhoti2023open,yi2023simple} 
propose to learn open-vocabulary semantic segmentation (OVSS) from web-crawled image-text paired data. 
These works can be categorized into two groups based on segmentation pipelines. Group-based methods, represented by GroupViT~\cite{xu2022groupvit}, cluster an image into distinct group tokens, which are averaged to align with text during training and independently matched to text at inference.
SegCLIP~\cite{luo2023segclip} arms CLIP~\cite{radford2021learning} by a plugged group module and introduces additional supervision including a reconstruction loss~\cite{he2022masked} and a superpixel-based KL loss~\cite{felzenszwalb2004efficient}.
PGSeg~\cite{zhang2023uncovering} leverages the prototypical knowledge extracted from both images and text to regularize group tokens. 
However, the image-group alignment gap between training and inference, compounded by under-segmentation issues arising from group unit, results in subpar zero-shot segmentation performance.

Another category, which we term pixel-wise methods~\cite{cha2023learning, mukhoti2023open, zhou2022extract}, delves into learning fine-grained alignment upon spatial image embeddings and performs pixel-wise classification at inference.
% conducts cross-modal alignment using spaital visual embeddings 
Rather than training from scratch, most works leverage the rich knowledge in large-scale visual-language pre-training~\cite{radford2021learning,jia2021scaling} and transfer the image-level alignment to finer alignment. MaskCLIP~\cite{zhou2022extract} introduces modifications to the last attention layer to obtain dense embeddings for segmentation. TCL~\cite{cha2023learning} incorporates a text grounding procedure to enhance the quality of region-text alignment. PACL~\cite{mukhoti2023open} proposes a modified compatibility function to implicitly learn the patch-level alignment. Nevertheless, the train-test region-pixel alignment gap, along with the over-segmentation issues caused by pixel unit hampers the zero-shot segmentation performance. In this study, we introduce an innovative framework designed to bridge the train-test alignment gap while enabling adaptive prediction for downstream segmentation.

\noindent\textbf{Open-vocabulary semantic segmentation.}  In addition to image-text pairs, \cite{liang2023open,zhang2023simple,xu2023side,xu2023open,zou2023generalized} also incorporates dense annotations during training to learn OVSS. They acquire segmentation capability from dense annotations and extend vocabulary through image-text supervision. Despite the impressive results, the reliance on dense annotations imposes limitations on the scalability and generalization of the methods. Therefore, our work focuses on learning OVSS using cheap and abundant web-crawled image-text pairs without dense annotations.

\section{Method}
\label{sect:method}

 In this section, we introduce the Multi-Grained Cross-modal Alignment (MGCA) framework for text-supervised open-vocabulary semantic segmentation.
 We first present the problem definition and target objective in \cref{sec:Problem definition and overall architecture}.  
 To achieve the objective, we ingeniously construct pseudo multi-grained semantic correspondence upon image-text pairs to facilitate the fine-grained alignment in \cref{sec:Align in all granularity}. 
 Lastly, we develop an adaptive and transferable semantic unit for downstream semantic segmentation tasks in \cref{sec:Transferable semantic units}.

\begin{figure*}[tb]
\centering
\includegraphics[scale=0.41]{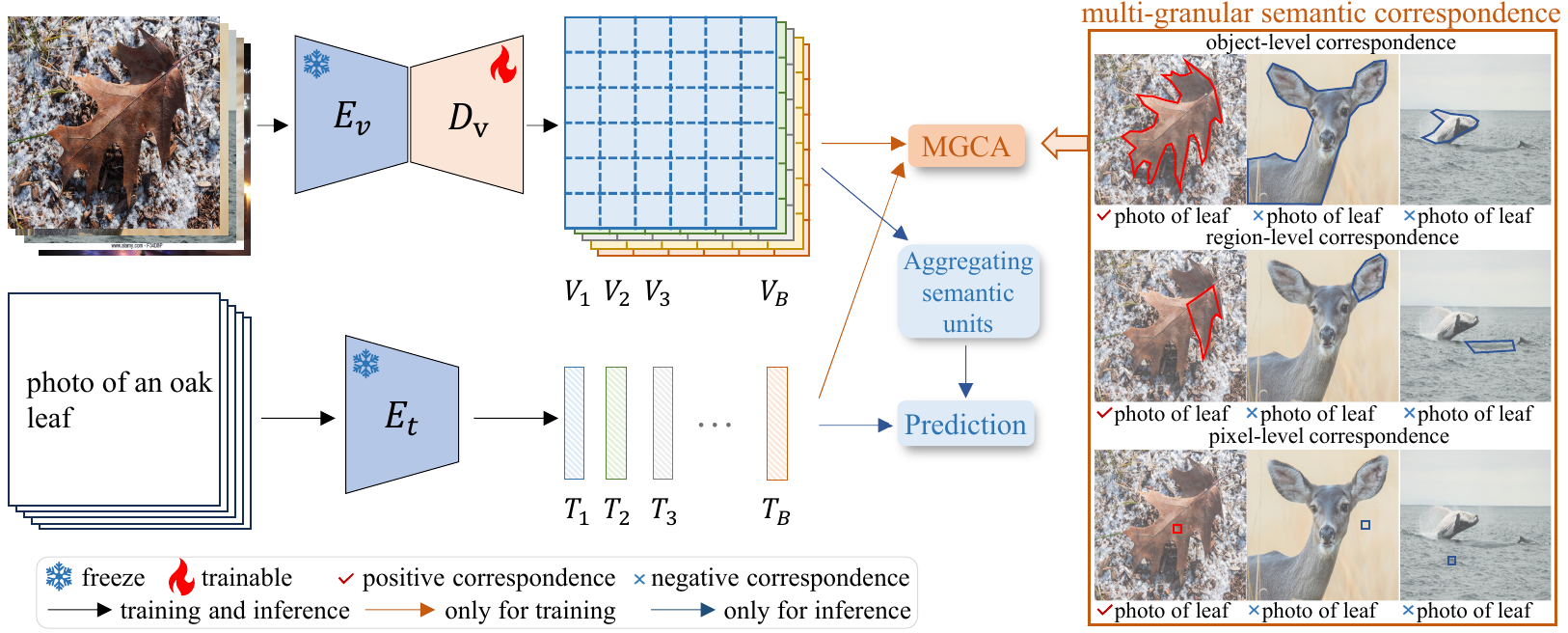}
\caption{The architecture and overall pipeline of our method. $E_v$ and $E_t$ are initialized with CLIP and frozen. We train the decoder $D_v$ only. 
As illustrated by the provided examples, Multi-Grained Cross-modal Alignment (MGCA) innovatively constructs object/region/pixel-level semantic correspondence, which enables the model to learn  
fine-grained alignment without dense annotations.
During inference, we discard MGCA and aggregate pixel embeddings into adaptive semantic units for predictions.}    
\vspace{-10pt}
\label{fig:overall}
\end{figure*}

\subsection{Problem Definition}
\label{sec:Problem definition and overall architecture}
We aim to learn fine-grained visual representations capable of generalizing to various downstream semantic segmentation tasks in a zero-shot manner purely with web-crawled image-text pairs.
An overview of the proposed method is presented in \cref{fig:overall}.
The architecture of our method consists of three modules: image encoder $E_v$, text encoder $E_t$, and a lightweight decoder $D_v$. $E_v$ and $E_t$ are initialized with the pre-trained CLIP~\cite{radford2021learning} weights and are both frozen to preserve and exploit the rich knowledge of the large-scale visual-language pre-training instead of training from scratch. 
We only train the lightweight decoder which converts visual features to dense pixel embeddings for fine-grained alignment.
Assume an input batch of image-text pairs $\{(x_i^I, x_i^T)\}_{i=1}^{B}$, where $x_i^I$ and $x_i^T$ denote an image and its paired caption, respectively.
For an image $x_i^I$, we first pass it through $E_v$ to extract the spatial image features, which exclude the global $[\text{CLS}]$ embedding. Then the spatial features are fed into the decoder $D_v$ to obtain the pixel embeddings $V_i$,  which are employed to align with text features $T_i \in \mathbb{R}^{C} $ extracted from the text encoder $E_t$.
We summarize the simple process as follows:
\begin{equation}
\label{eq: extract features}
    V_i = D_v(E_v(x_i^I)),\qquad
    T_i = E_t(x_i^T)
\end{equation}
where $V_i = \{V_i^p, p=1,...,L \} \in \mathbb{R}^{L\times C}$ represents the sequence of the final pixel embeddings. $L$ and $C$ indicate the number of pixels and the embedding dimension size. 
Formally, the similarity between $p$th pixel embeddings of $i$th image $V_i^p$ and any text embeddings $T_j$ can be computed as the following dot production:
\begin{equation}
\label{eq: similarity}
    S_{ij}^p = V_i^p \cdot T_j
\end{equation}

The pixel-to-text similarity matrix $S_{ij}$ reflects the degree of semantic correlation between dense visual representations and text embeddings. The objective of open-vocabulary semantic segmentation is to learn fine-grained cross-modal alignment.
In this context, \textcolor{magenta}{the fine-grained alignment asks to improve the similarity $S_{ij}^p$ when $i$ equals $j$ and the $pth$ pixel falls into the text-described region otherwise decrease the similarity}. However, the absence of matching between pixels and texts makes learning fine-grained alignment challenging.

\begin{figure*}[tb]
\centering
\includegraphics[scale=0.41]{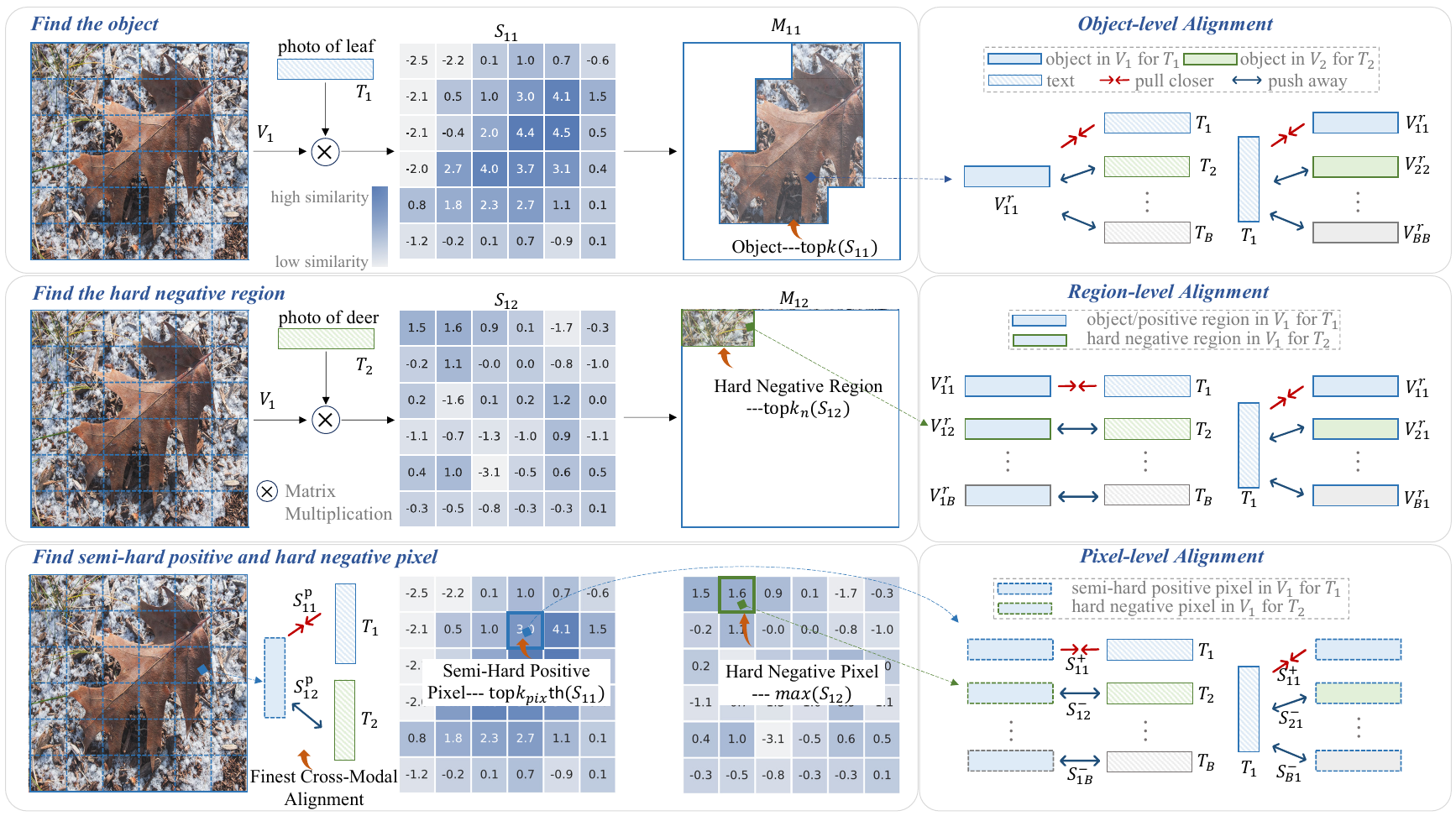}%%%%%%%%%%%%%%%%scale=缩小比例，或者用width=2in
\caption{Illustration of the proposed Multi-Grained Cross-modal Alignment (MGCA). Based on the pixel-to-text similarity matrix $S_{ij}$, we identify informative positive and negative pairs for object-, region- and pixel-level contrastive learning. 
}     
\label{fig:mgca}
\vspace{-10pt}
\end{figure*}

\subsection{Multi-Grained Cross-Modal Alignment}
\label{sec:Align in all granularity}
Digging into web-crawled data, we ingeniously construct pseudo semantic correspondence at object-, region- and pixel-level to bridge the granularity gap and enhance the fine-grained alignment.
The pipeline of MGCA is presented in \cref{fig:mgca}.

\textbf{Align at object-level.}
The most straightforward way to perform fine-grained alignment is to transfer image-level supervision to object-level, \ie, locate the text-described object in each image and align it with text in the latent space.
Given the similarity matrix ${S_{ii}} \in \mathbb{R}^{L}$ of paired image-text, we first calculate the probability of each pixel falling into the text-described object by applying the softmax function across pixels:
${\mathit{Prob}_{ii}} = \text{softmax}(S_{ii}) \in \mathbb{R}^{L}$. 
Then, pixels with top-$k$ similarity within the image are considered to form the target object, and its mask can be computed as follows:
\begin{equation}
\label{eq:obj mask}
M_{ii}^p = 
\begin{cases}
1, & \mathit{Prob}_{ii}^p \geq \text{top-}k(\mathit{Prob}_{ii})  \\
0, & \mathit{Prob}_{ii}^p \textless \text{top-}k(\mathit{Prob}_{ii})\\
\end{cases} \qquad
p = 0, 1, ..., L
\end{equation}
where $k$ represents the maximum area of the object. 
To further attend to the reliable pixels and suppress the adverse impact of ambiguous pixels with low similarity within the target, we weight the pixel embeddings with ${\mathit{Prob}_{ii}}$ to get the object-level representation as follows:
\begin{equation}
\label{eq:obj feature}
    V_{ii}^r = (V_i)^\top({\mathit{Prob}_{ii}} \odot M_{ii})
\end{equation}
where ${V_{ii}^r} \in \mathbb{R}^{C}$ and $\odot$ denotes element-wise multiplication.
The objective of object-level cross-modal alignment can be formulated as follows:
\begin{equation}
\label{Eq:obj}
    \mathcal L^{obj} = -\dfrac{1}{B}\sum\limits_{i=1}^B(\text{log}\dfrac{\text{exp}(s(V_{ii}^r, {T_i})/\tau)}{\sum_{j=1}^{B}\text{exp}(s(V_{ii}^r, {T_j})/\tau)} + 
    \text{log}\dfrac{\text{exp}(s({T_i} , V_{ii}^r)/\tau)}{\sum_{j=1}^{B}\text{exp}(s({T_i}, V_{jj}^r)/\tau)})
\end{equation}
where $s(x,y)=\dfrac{x\cdot y}{\Vert x \Vert_2 \Vert y \Vert_2}$ is the cosine similarity. $B$ is the batch size and  $\tau$ is the temperature hyperparameter.

\textbf{Align at region-level.} The object-level alignment is insufficient to establish semantic correspondence between non-object regions and texts. Therefore, we further integrate the region-level alignment, which encourages the model to map the positive region-text pairs nearby while mapping all negative pairs far apart in the latent space. 

The object-text pairs above naturally provide proper positive region correspondences.
Any region in an image with its unpaired text forms a negative region-text pair.
However, redundant negatives impede the computational efficiency and the learning of discriminative representations.
In this respect, we propose to boost region-level alignment with an efficient hard negative mining strategy.
Concretely, we establish negatives by selecting the most relevant region in an image to the unpaired text using \cref{eq:obj feature}. 
Recall that $i$, $j$ in  $S_{ij}$ denote the sequence order of image and text in a batch, respectively. Hence,  $V_{ij}^r$ seamlessly represents the embedding of the hard negative region $M_{ij}$ when $i \neq j$. 
We delineate the objective of region-level cross-modal alignment as follows:
\begin{equation}
    \mathcal L^{reg} = -\dfrac{1}{B}\sum\limits_{i=1}^B(\text{log}\dfrac{\text{exp}(s(V_{ii}^r,  {T_i})/\tau)}{\sum_{j=1}^{B}\text{exp}(s(V_{ij}^r,{T_j})/\tau)}
    + \text{log}\dfrac{\text{exp}(s({T_i}, V_{ii}^r) /\tau)}{\sum_{j=1}^{B}\text{exp}(s({T_i}, V_{ji}^r)/\tau)})
\end{equation}

\textbf{Align at pixel-level.} 
Pixel-level cross-modal alignment serves as a critical bridge to tackle the alignment granularity gap and achieve high-quality segmentation masks, as it fulfills the granularity demands of segmentation.
The objective of pixel-level alignment is to enhance the similarity between positive pixel-level correspondence and reduce the similarity between unmatched pairs.

Positive pixel-text pairs lie within the text-described objects, \ie, $i=j$ and $M_{ij}^p = 1$. The remaining pairs are regarded as negatives.
However, the pixel-level pairs present in a batch, especially the negatives, are severely redundant. Training with all candidates incurs high computational cost and the redundant easy ones would undermine the informative gradient information from hard negatives, hindering the learning of discriminative representations. 
Additionally, the hardest positive pixel within the text-described object may be negative indeed due to the lack of groundtruth masks, which would harm the performance.

To this end, 
we propose to collect hard negatives and semi-hard positives to facilitate pixel-level contrastive learning.
Selection of negative correspondence can be done by simply picking the pixel with the highest similarity in an image to its unpaired text.
To establish the hardness-trustworthiness trade-off positive correspondence, pixels with the top-$k_{pix}$th similarity in the objects are selected as semi-hard positives.
We maximize the similarity of positive pixel-text pairs versus negatives via contrastive learning as follows:

\begin{equation}
\label{eq:patch loss}
    \begin{aligned}
    \mathcal L^{pix} = -\dfrac{1}{B}\sum\limits_{i=1}^B &(\text{log}\dfrac{\text{exp}(S_{ii}^{+}/\tau)}{\text{exp}(\dfrac{S_{ii}^{+}}{\tau}) + \sum\limits_{j\neq i}^{B} \text{exp}(\dfrac{{S_{ij}^{-}}}{\tau})}
    +\text{log}\dfrac{\text{exp}({S_{ii}^{+}}/\tau)}{\text{exp}(\dfrac{{S_{ii}^{+}}}{\tau}) + \sum\limits_{j\neq i}^{B}  \text{exp}(\dfrac{{S_{ji}^-}}{\tau})})\\
    S_{ii}^{+} &= \text{top-}k_{pix}\text{th}(S_{ii}/f), \qquad
    S_{ij}^{-} = \max(S_{ij}/f)~s.t.~j \neq i
    \end{aligned}
\end{equation}
where $S_{ii}^{+}$, $S_{ij}^{-}$ represent the similarity of semi-hard positives and hard negatives, respectively. In practice, we directly extract the selected pixel-to-text similarity from $S_{ij}$ in \cref{eq: similarity} instead of recalculating their cosine similarity, and a scale factor $f$ is employed here to ensure the numerical stability of exponential.

\textbf{Overall objective.}~We train the model by jointly optimizing object-, region- and pixel-level cross-modal alignment. 
The overall objective can be represented as follows:
\begin{equation}
\label{eq:final loss}
\mathcal L = \mathcal L^{obj}  + \mathcal L^{reg} + \mathcal L^{pix}
\end{equation}

\subsection{Adaptive semantic unit}
\label{sec:Transferable semantic units}
To generalize to various downstream segmentation tasks, it is critical to employ an adaptive and transferable semantic unit at inference.
The semantic unit refers to the basic visual element processed by the recognition module, which means that pixels within a unit share the same category. However, it's noteworthy that existing units impose limitations on the segmentation performance.

Group-based methods recognize each group unit, which is formed by hierarchical grouping of visual tokens and results in coarse object-level representations.
However, owing to the limited initial global cluster centers and the absence of explicit group-level supervision, group unit suffers from imprecise grouping and under-segmentation when applied to downstream tasks as shown in \cref{fig:result}.  
In contrast, pixel-wise methods identify each pixel independently and disregard the semantic-level visual coherence. Due to the lack of sufficient pixel-level supervision, individual pixel embedding may not carry adequate semantics and be susceptible to noise. 
As a result, inconsistent predictions appear among the pixels within regions with the same semantics, leading to over-segmentation as depicted in \cref{fig:result}. 

Based on above analysis, rather than utilizing existing units directly, we develop an adaptive semantic unit that cleverly harnesses their advantages. 
The core idea is to \textcolor{magenta}{finely aggregate} pixels with relevant semantics in the final visual embeddings based on pixel affinity. 
\textbf{Firstly}, we conduct uniform sampling on the dense pixel embeddings to acquire a collection of meta-points $\mathcal{P}$.
\textbf{Secondly}, we assign each pixel to its most similar meta-point in $\mathcal{P}$, thus obtaining a set of pixels for each meta-point. \textbf{Lastly}, each set of pixels is averaged to compute its semantic embedding, which is employed to calculate the most similar text as the prediction result.

\begin{figure*}[tb]
\centering
\includegraphics[scale=0.41]{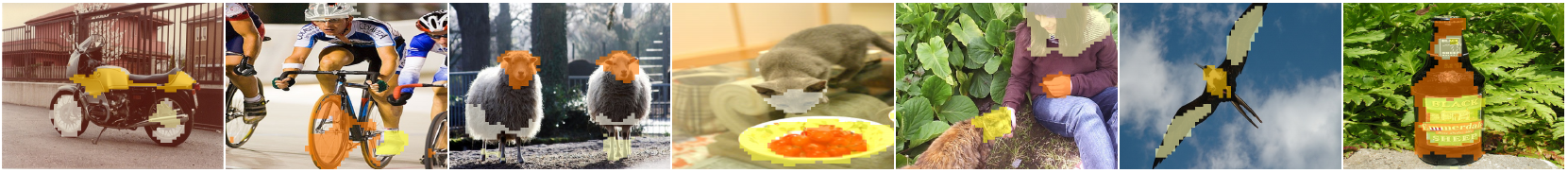}%%%%%%%%%%%%%%%%scale=缩小比例，或者用width=2in
\caption{Examples of our semantic units. Pixels with the same color belong to the same unit. Our semantic units align with part-level representations, such as the wheel hub in the first image and the hat and shoes in the second image.} 
\vspace{-10pt}
\label{fig:semantic unit}
\end{figure*}

The proposed semantic unit dynamically exploits the generalizable visual coherence and finely aggregates semantically relevant pixels, forming valid part-level representations as illustrated in \cref{fig:semantic unit}. Unlike group-based methods rely on limited group tokens and global initial centers, our part-level unit benefits from adequate local sampling meta-points, ensuring a fine and precise aggregation of pixels. Moreover, the average aggregation effectively enhances the joint features across pixels and mitigates the adverse impact of inconstant noise with minimal computational overhead. 
The adaptive unit enables the feasible utilization of the learned multi-grained alignment capabilities and alleviates the under- and over-segmentation problems.

\section{Experiments}
\label{sec:experiments}
\subsection{Implementation details}
\textbf{Architecture}
We adopt the visual encoder and text encoder from CLIP~\cite{radford2021learning} ViT-B/16~\cite{dosovitskiy2020image}. The input image size is 224 $\times$ 224 and the patch size is 16 $\times$ 16 during the training stage. We extract spatial visual features from the image encoder following ~\cite{zhou2022extract, cha2023learning} that modifies the last attention layer. The learnable decoder, with a total of \textcolor{magenta}{4.72}M parameters, consists of an upsampling interpolation with a scale factor of 2 and two gated convolution blocks from ~\cite{cha2023learning}.

\noindent\textbf{Training} We train our models purely on CC3M~\cite{sharma2018conceptual}. Discounting expired links, a total of 2.56 million image-text pairs are actually downloaded. To enrich positive correspondence, we extract nouns from each caption and use CLIP prompts to generate captions with these nouns as GroupViT~\cite{xu2022groupvit}. When forming a batch, we randomly select one of the captions for each image from either its original caption or the generated captions. The models are trained on 4 RTX 3090 GPUs with a global batch size of 1024. The learning rate is 1e-4 and decayed via the cosine annealing schedule~\cite{loshchilovstochastic}. We train the models for 15 epochs, with a linear warmup for the first 2 epochs. Adam optimizer~\cite{kinga2015method} with a weight decay of 0.05 is used. Following the practice in contrastive learning~\cite{he2020momentum}, the temperature hyperparameter $\tau$ is 0.07. And inspired by~\cite{vaswani2017attention}, we set the scale factor in \cref{eq:patch loss} as $f = 5\times \sqrt{C}$, $C$ is the embedding dimension size.

\noindent\textbf{Hyperparameters}
The area $k \in [0,1]$ in \cref{eq:obj mask} for the positive object is 0.4, which corresponds to the average area of text-described regions in CC3M~\cite{sharma2018conceptual}.
For negative region-level pairs, we denote the area as $k_{n}$ and set it to 0.05. 
The $k_{pix}\in[0, 1]$ in \cref{eq:patch loss} for semi-hard positives is 0.06. 

\noindent\textbf{Inference} Following previous works~\cite{xu2022groupvit, cha2023learning}, we resize the input image to have a shorter size of 448 and adopt a sliding window strategy during inference. 
Pixel embeddings from the decoder are aggregated into 36 semantic units per image, that is, the number of meta-points is 36.
 Within each unit, the pixel embeddings are averaged to calculate their corresponding semantic embedding, which is then utilized to calculate the most similar text as the prediction result.

\begin{table*}[t]
\centering
\caption{Zero-shot segmentation performance comparison on datasets with a background class. All methods are reported without post-processing. mIoU~(\%) metric is used in every experiment. Abbreviations of benchmarks, from left to right: Pascal VOC, Pascal Context, COCO-Object. Results above the dotted line are taken from papers that may not adhere to the unified evaluation protocol. * represents the results are trained on the downloaded CC3M. We highlight the \textbf{best} and \underline{second-best} results.}
\label{tab1}
\resizebox{0.9\linewidth}{!}{
\begin{tabular}{c|c|ccc|c}
\hline
\textbf{Method} & Pretrain Data & VOC & Context & COCO & Avg.\\
\hline
ViL-Seg~$_\mathrm{\textcolor{gray}{[CVPR23]}}$~\cite{liu2022open} & CC12M~\cite{changpinyo2021conceptual} & 37.3 & 18.9 & 18.1 & 24.8\\
ViewCo~$_\mathrm{\textcolor{gray}{[ICLR23]}}$~\cite{ren2022viewco} & CC12M+YFCC14M~\cite{thomee2016yfcc100m} & 52.4 & 23.0 & 23.5 & 33.0\\
MixReorg~$_\mathrm{\textcolor{gray}{[ICCV23]}}$~\cite{cai2023mixreorg} & CC12M & 47.9 & 23.9 & 21.3 & 31.0\\
CoCu~$_\mathrm{\textcolor{gray}{[NeurIPS23]}}$~\cite{xing2023rewrite}  & CC3M\&12M+YFCC14M& 49.7 & 22.8 & 22.0 & 31.5\\
\cdashline{1-6}[1pt/1pt]
GroupViT~$_\mathrm{\textcolor{gray}{[CVPR22]}}$~\cite{xu2022groupvit} & CC3M\&12M+YFCC14M & 52.3 & 22.4 & 24.3 & 33.0\\
GroupViT~$_\mathrm{\textcolor{gray}{[CVPR22]}}$~\cite{xu2022groupvit} & CC3M\&12M+RedCaps12M~\cite{desai2021redcaps} & 50.8 & 23.7 & 27.5 & 34.0\\
PGSeg~$_\mathrm{\textcolor{gray}{[NeurIPS23]}}$~\cite{zhang2023uncovering}  & CC12M + RedCaps12M & \textbf{53.2} & 23.8 & 28.7 & 35.2 \\
SegCLIP~$_\mathrm{\textcolor{gray}{[ICML23]}}$~\cite{luo2023segclip} & CC3M+COCO~Caption~\cite{lin2014microsoft}  & 52.6 & 24.7 & 26.5 & 34.6\\
MaskCLIP~$_\mathrm{\textcolor{gray}{[ECCV22]}}$~\cite{zhou2022extract}  & -  & 38.8 & 23.6 & 20.6 & 27.7\\
$\text{TCL}^*$~$_\mathrm{\textcolor{gray}{[CVPR23]}}$~\cite{cha2023learning} & CC3M  & 51.4 & \underline{25.1} & 28.8 & 35.1\\
TCL~$_\mathrm{\textcolor{gray}{[CVPR23]}}$~\cite{cha2023learning} & CC3M\&12M  & 51.2 & 24.3 & \underline{30.4} & \underline{35.3}\\
\rowcolor{blue!10}MGCA~(Ours) & CC3M & \underline{53.1} & \textbf{31.3} & \textbf{31.9} & \textbf{38.8}\\

\hline
\end{tabular}
}                
\end{table*}

\begin{table*}[tb]
\centering
\caption{Zero-shot segmentation performance comparison on datasets without background class. Abbreviations of benchmarks, from left to right: Pascal VOC20, Pascal Context59, COCO-Stuff, Cityscapes, Ade20k.}
\label{tab2}
\resizebox{0.9\linewidth}{!}{
\begin{tabular}{c|c|ccccc|c}
\hline
\textbf{Method} & Pretrain Data & VOC20 & C59 & STUFF & CITY & ADE & Avg.\\
\hline
CoCu~$_\mathrm{\textcolor{gray}{[NeurIPS23]}}$~\cite{xing2023rewrite} & CC3M\&12M+YFCC14M& - & - & 14.9 & 21.9 & 12.0 & -\\
\cdashline{1-8}[1pt/1pt]
GroupViT~$_\mathrm{\textcolor{gray}{[CVPR22]}}$~\cite{xu2022groupvit} & CC3M\&12M+YFCC14M & 74.1 & 20.8 & 12.6 & 6.9 & 8.7 & 24.6 \\
GroupViT~$_\mathrm{\textcolor{gray}{[CVPR22]}}$~\cite{xu2022groupvit} & CC3M\&12M+RedCaps12M & 79.7 & 23.4 & 15.3 & 11.1 & 9.2 & 27.7\\
PGSeg~$_\mathrm{\textcolor{gray}{[NeurIPS23]}}$~\cite{zhang2023uncovering}  & CC12M + RedCaps12M & 79.6 & 23.4 & 15.3 & 12.5 & 9.1 & 28.0\\
SegCLIP~$_\mathrm{\textcolor{gray}{[ICML23]}}$~\cite{luo2023segclip} & CC3M+COCO~Caption  & 76.8 & 25.6 & 16.5 & 11.0 & 8.7 & 27.7\\
MaskCLIP~$_\mathrm{\textcolor{gray}{[ECCV22]}}$~\cite{zhou2022extract} & -  & 74.9 & 26.4 & 16.4 & 12.6 & 9.8 & 28.0\\
$\text{TCL}^*$~$_\mathrm{\textcolor{gray}{[CVPR23]}}$~\cite{cha2023learning} & CC3M  & 76.8 & 28.4 & 18.7 &  20.1 & 14.3 & 31.7\\
TCL~$_\mathrm{\textcolor{gray}{[CVPR23]}}$~\cite{cha2023learning} & CC3M\&12M  & \underline{77.5} & \underline{30.3} & \underline{19.6} & \underline{23.1} & \underline{14.9} & \underline{33.1} \\
\rowcolor{blue!10}MGCA~(Ours)& CC3M & \textbf{81.4} & \textbf{33.7} & \textbf{22.0}& \textbf{24.0} &\textbf{16.4}& \textbf{35.5} \\
\hline

\end{tabular}
}
\vspace{-10pt}
\end{table*}

\subsection{Zero-shot Transfer to Semantic Segmentation} 
\textbf{Evaluation protocol} We follow the unified evaluation protocol in TCL~\cite{cha2023learning} to evaluate our method on widely used 8 benchmark datasets, categorized into two groups: $(i)$ with a background class including Pascal VOC~(21 classes)~\cite{everingham2010pascal}, Pascal Context~(60 classes)~\cite{mottaghi2014role}, COCO-Object~(81 classes)~\cite{caesar2018coco}, $(ii)$ without background class including Pascal VOC20~(20 classes)~\cite{everingham2010pascal}, Pascal Context59~(59 classes)~\cite{mottaghi2014role}, COCO-Stuff~(171 classes)~\cite{caesar2018coco},  ADE20k~(150 classes)~\cite{zhou2019semantic} and City-scapes (19 classes)~\cite{cordts2016cityscapes}. For the first category, we evaluate the foreground classes by thresholding the softmax-normalized-similarity between the pixel embeddings and text features following~\cite{xu2022groupvit}. For the other datasets, we evaluate both object and stuff classes. The unified class names from MMSegmentation~\cite{contributors2020mmsegmentation} is used without class name based tricks. We employ the text prompts following TCL~\cite{cha2023learning} without any test-time augmentation. The post-processing methods such as dense CRF~\cite{krahenbuhl2011efficient} and PAMR~\cite{araslanov2020single} are not used for a fair comparison.

\noindent\textbf{Comparison with the state-of-the-art}
We quantitatively compare our method with the state-of-the-art methods on the two groups of benchmarks. 
We summarize the performance reported in the papers (which may not adhere to the unified evaluation protocol) for methods that have not released the pretrained weights. 
And most group-based methods are compared only in \cref{tab1} such as ViL-Seg~\cite{liu2022open}, ViewCo~\cite{ren2022viewco} and  MixReorg~\cite{cai2023mixreorg}  since they tend to focus on evaluating and presenting the performance on object-oriented datasets, \ie, the datasets with a background class.
For the methods that have released their pretrained weights including GroupViT~\cite{xu2022groupvit}, PGSeg~\cite{zhang2023uncovering}, SegCLIP~\cite{luo2023segclip}, MaskCLIP~\cite{zhou2022extract} and TCL~\cite{cha2023learning}, we evaluate and present their performance across all benchmarks following the unified protocol. In addition, we train the current best-performing method TCL~\cite{cha2023learning} on the downloaded CC3M using their official code to provide a fair comparison.

\cref{tab1} illustrates the comparison results on the object-oriented datasets.
% that contain a ``background'' class. 
Our MGCA achieves substantial advancements over preceding state-of-the-art methods while utilizing a reduced amount of training data. Concretely, we exceed the current best-performing method TCL~\cite{cha2023learning} by an average of 3.5\% mIoU. 
\cref{tab2} summarizes the performance comparison results on the datasets that compromise both object and stuff classes. It further demonstrates MGCA attains noticeable improvements across various downstream tasks with diverse scenes and categories.
Compared to high-performing TCL~\cite{cha2023learning} which involves a total of 15M training samples, MGCA exhibits superior performance with a 2.4\%  increase in average mIoU, despite utilizing only 3M training samples. More quantitative comparison results are provided in Appendix.

\begin{figure*}[tb]
\centering
\includegraphics[scale=0.4]{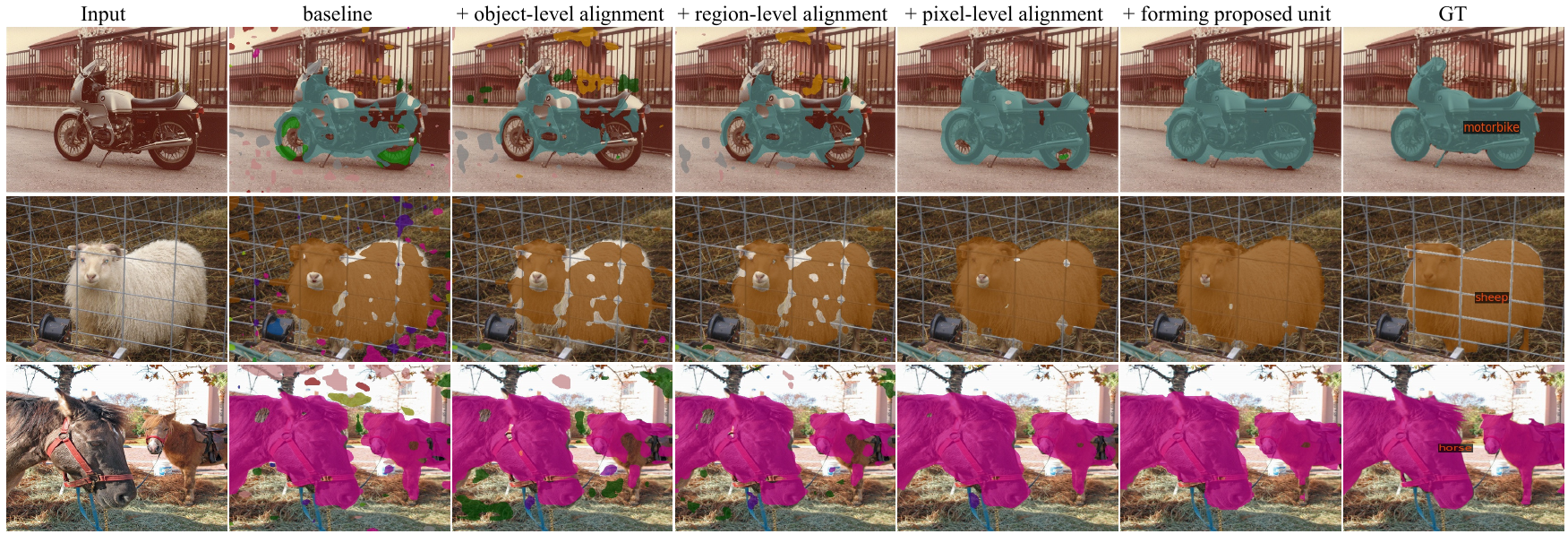}%%%%%%%%%%%%%%%%scale=缩小比例，或者用width=2in
\caption{Visualization of the impact of each module. The baseline corresponds to the first row in \cref{tab: grained} which directly employs CLIP for dense prediction. We progressively integrate object-, region- and pixel-level alignment modules, along with the proposed unit, into the model to qualitatively demonstrate the impact of each module. }     
\label{fig:ablation}
\vspace{-10pt}
\end{figure*}
Notably, our method achieves these results without using any dense annotations, and without introducing any additional supervision (\eg, superpixel), any regularization or any post-processing techniques. 
We will release the code with the pretrained checkpoint upon acceptance. \textcolor{blue}{Limitations and future works} are discussed in the Appendix.

\begin{table*}[tb]
\centering
\caption{Ablation studies of alignment granularity. Pixel units are employed to clearly illustrate the impact of fine-grained alignment. We present the absolute improvement in mIoU (\%) for each setting relative to the baseline. Settings in our method are highlighted in \colorbox{blue!10}{blue}.}
\label{tab: grained}
\resizebox{0.95\linewidth}{!}{
\begin{tabular}{ccc|ccc|c|ccccc|c}
\hline
object & region & pixel & \multicolumn{4}{c|}{with background class} & \multicolumn{6}{c}{without background class} \\
\cline{4-13}
level & level & level & VOC & Context & COCO & Avg. & VOC20 & C59 & STUFF & CITY & ADE & Avg. \\
\hline
\textcolor[rgb]{0.75,0.75,0.75}{\ding{55}} & \textcolor[rgb]{0.75,0.75,0.75}{\ding{55}} & \textcolor[rgb]{0.75,0.75,0.75}{\ding{55}} &
 25.9 & 17.1 & 20.0 & 21.0 & 40.1 & 18.3 & 10.0 & 17.2 & 8.2 & 18.8 \\
 \cdashline{1-13}[1pt/1pt]
\checkmark &  \textcolor[rgb]{0.75,0.75,0.75}{\ding{55}} &  \textcolor[rgb]{0.75,0.75,0.75}{\ding{55}} & 42.6 & 27.8 & 24.3 & 31.6~(\textcolor[rgb]{0,0,1}{10.6~$\uparrow$})  & 66.5 & 28.7 & 17.5 & 23.7 & 12.6 & 29.8~(\textcolor[rgb]{0,0,1}{11.0~$\uparrow$}) \\
\textcolor[rgb]{0.75,0.75,0.75}{\ding{55}} & \checkmark &  \textcolor[rgb]{0.75,0.75,0.75}{\ding{55}} & 41.6 & 22.8 & 24.6 & 29.7~(\textcolor[rgb]{0,0,1}{~8.7~$\uparrow$})~& 63.0 & 24.6 & 15.1 & 22.1 & 11.6 & 27.3~(\textcolor[rgb]{0,0,1}{~8.5~$\uparrow$})~ \\
\textcolor[rgb]{0.75,0.75,0.75}{\ding{55}} & \textcolor[rgb]{0.75,0.75,0.75}{\ding{55}} & \checkmark & 50.4 & 29.2 & 29.3 & 36.3 (\textcolor[rgb]{0,0,1}{15.3~$\uparrow$}) & 78.2& 31.1 & 20.1 & 20.9 &15.3 & 33.1~(\textcolor[rgb]{0,0,1}{14.3~$\uparrow$}) \\
 \cdashline{1-13}[1pt/1pt]
 \checkmark &  \textcolor[rgb]{0.75,0.75,0.75}{\ding{55}} &  \textcolor[rgb]{0.75,0.75,0.75}{\ding{55}} & 42.6 & 27.8 & 24.3 & 31.6~(\textcolor[rgb]{0,0,1}{10.6~$\uparrow$})  & 66.5 & 28.7 & 17.5 & 23.7 & 12.6 & 29.8~(\textcolor[rgb]{0,0,1}{11.0~$\uparrow$}) \\
\checkmark & \checkmark  &  \textcolor[rgb]{0.75,0.75,0.75}{\ding{55}}& 44.1 & 28.9 & 26.2 & 33.1 (\textcolor[rgb]{0,0,1}{12.1~$\uparrow$})& 71.6 & 30.2 & 19.2 & 24.9 & 14.5 & 32.1~(\textcolor[rgb]{0,0,1}{13.3~$\uparrow$}) \\
\rowcolor{blue!10}\checkmark & \checkmark  & \checkmark & 51.1 & 30.0 & 30.3 & 37.1 (\textcolor[rgb]{0,0,1}{16.1~$\uparrow$}) &77.5  & 31.8 & 20.5 & 23.2 & 15.6 & 33.7 (\textcolor[rgb]{0,0,1}{14.9~$\uparrow$})\\
\hline

\end{tabular}
}
\end{table*}

\begin{figure}[htbp]
\centering
\begin{minipage}[t]{0.492\textwidth}
\centering
\includegraphics[width=0.7\linewidth]{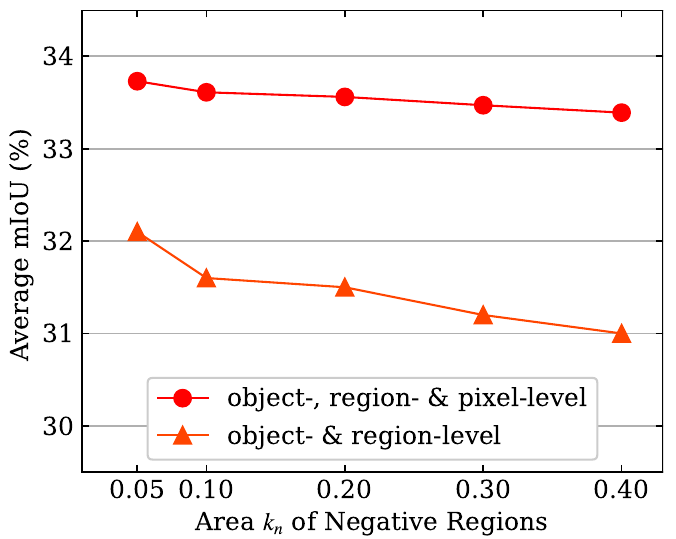}
\caption{Impact of $k_{n}$. $k_{n}$ decides the area of regions for negative region-level correspondence. We present the performance on datasets without background class. }
\label{fig:knobj}%文中引用该图片代号
\end{minipage}
\hspace{.05in}
\begin{minipage}[t]{0.45\textwidth}
\centering
\includegraphics[width=0.92\linewidth]{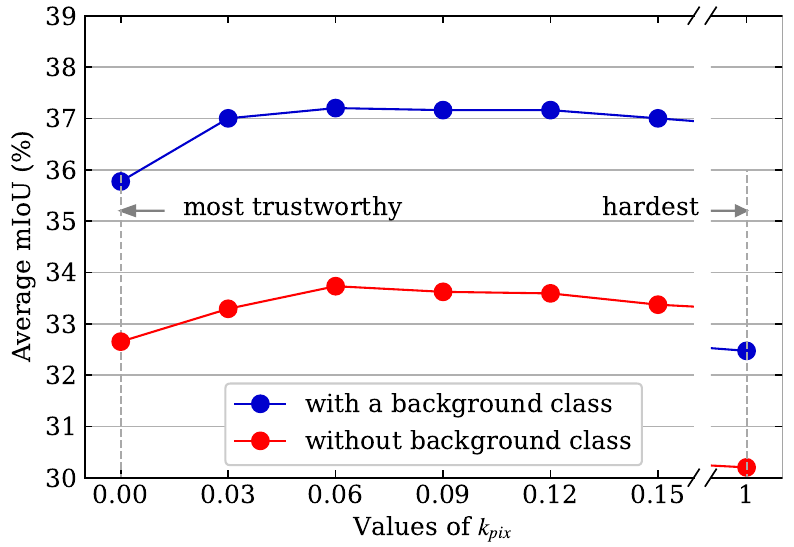}
\caption{Impact of $k_{pix}$. $k_{pix}$ controls the hardness-trustworthiness trade-off for  positive pairs in pixel-level alignment.}
\label{fig:kpat}%文中引用该图片代号
\end{minipage}
\vspace{-15pt}
\end{figure}

\subsection{Ablation Studies}
\label{sec:Ablation}
In this section, we conduct ablation studies to evaluate the effects of core components of the proposed method. 

\noindent\textbf{Impact of alignment granularity.} 
\cref{tab: grained} presents the ablation studies on the alignment granularity. 
To better understand and assess the impact of fine-grained alignment on the segmentation performance, we adopt pixel as semantic unit here in accordance with previous pixel-wise methods.
The first row serves as the baseline that directly calculates the most similar text to each spatial embedding from the CLIP image encoder as the prediction. 
The quantitative results demonstrate:
(1) Each alignment granularity is proven to be beneficial for dense prediction,
\eg, the object-level alignment enhances the baseline by 10.6\% mIoU on object-oriented datasets and region-level alignment delivers an 8.5\% improvement on the datasets compromise both object and stuff classes. 
(2) The finest pixel-level alignment brings the most significant improvements. In particular, pixel-level alignment alone achieves remarkable improvements of 15.3\% and 14.3\% on the two groups of benchmarks, respectively. 
(3) Combining the three-level granularity alignment yields the best performance,  resulting in an average improvement of 16.1\% and 14.9\%  mIoU compared to the baseline.

We qualitatively explore the effect of alignment granularity by incrementally incorporating object-, region- and pixel-level alignment into the baseline. As illustrated in \cref{fig:ablation}, object- and region-level alignment effectively enhance the identification of salient regions and mitigates the occurrence of false cross-modal correspondences, \eg, the noisy patches in the background. And incorporating the pixel-level alignment significantly improves the completeness and accuracy of segmentation masks.

\begin{table*}[tb]
\centering
\caption{Ablation studies of semantic unit and the number of meta-points. 
% The number of semantic units per image is calculated as dividing the patch length by the sampling interval (>0). 
The first line employs the pixel unit, where the number of units is equal to the total number of pixel embeddings.}
\label{tab:meta points}
\resizebox{0.9\linewidth}{!}{
\begin{tabular}{c|c|cccc|cccccc}
\hline
 semantic & \# meta-& \multicolumn{4}{c|}{with background class} & \multicolumn{6}{c}{without background class} \\
 unit & points  & VOC & Context & COCO & Avg. & VOC20 & C59 & STUFF & CITY & ADE & Avg. \\
\hline
 pixel & 3136  & 51.1 & 30.0 & 30.3 & 37.1 & 77.5 & 31.8 & 20.5 & 23.2 &15.6 & 33.7 \\
\cdashline{1-12}[1pt/1pt]
\multirow{7}*{Ours}& 
121 & 52.3 & 30.9 & 31.3 & 38.2 & 80.0 & 33.1 & 21.5 & 23.9 & 16.3 & 35.0 \\
 & 81 & 52.5 & 31.1 & 31.5 & 38.4 & 80.4 & 33.3 & 21.7 & 23.9 & 16.4 & 35.1 \\
& 64 & 52.7 & 31.1 & 31.6 & 38.5 & 80.6 & 33.5 & 21.8 & 24.0 & 16.4 & 35.3 \\
& 49 & 52.8 & 31.2 & 31.8 & 38.6 & 81.0 & 33.6 & 22.0 & 23.9 & 16.4 & 35.4 \\
& \cellcolor{blue!10}36 & \cellcolor{blue!10}53.1 &\cellcolor{blue!10}31.3 & \cellcolor{blue!10}31.9 & \cellcolor{blue!10}38.8 &\cellcolor{blue!10}81.4 &\cellcolor{blue!10}33.7 & \cellcolor{blue!10}22.0 & \cellcolor{blue!10}24.0 & \cellcolor{blue!10}16.4 & \cellcolor{blue!10}35.5 \\
& 25 & 53.3 & 31.3 & 31.9 & 38.8 & 81.9 & 33.8 & 22.2 & 23.3$_\downarrow$ & 16.2$_\downarrow$ & 35.5  \\
& 16 & 53.7 & 31.1 & 31.7 & 38.8 & 82.2 & 33.8 & 22.1 & 22.6 & 15.9 & 35.3  \\
\hline
\end{tabular}
}
\vspace{-10pt}
\end{table*}
\noindent\textbf{Impact of $k_n$}. 
Recall that we calculate the mask of hard negative region by selecting pixels with top-$k_n$ similarity to the unpaired text.
\cref{fig:knobj} explores the influence of the area $k_{n}$ for negative region-level correspondence on datasets without background class.
We observe that alterations in ${k_n}$ exert little effect on model performance. Therefore we also present the results excluding pixel-level alignment to clearly access the impact of ${k_n}$.   
As expected, smaller $k_{n}$ compels the model to distinguish more challenging regions and promotes finer alignment which benefits the dense predictions as shown in \cref{fig:knobj}.

\noindent\textbf{Impact of $k_{pix}$}. \cref{fig:kpat} quantifies the effect of $k_{pix}$  which controls the hardness and trustworthiness of positive pixel-text pairs. 
During training, an image is assumed to contain $k \times L$ positive pixels, where $k=0.4$ as mentioned above, and $L$ is the total number of pixel embeddings.
Bigger $k_{pix}$ signifies harder positives which may result in false positive predictions, whereas smaller $k_{pix}$ is likely to cause false negative predictions. The results presented in \cref{fig:kpat} demonstrate the superiority of semi-hard positives $k_{pix}=0.06$ compared to the hardest positives $k_{pix}=1.0$ and the most trustworthy positives $k_{pix}=0$. The impact of hard mining strategy for negative pixel-level correspondence is provided in Appendix.

\noindent\textbf{Impact of the semantic unit and the number of meta-points.}
\cref{tab:meta points} reports the performance on all benchmarks with regard to the semantic units. 
The first row serves as the baseline, aligning with pixel-wise methods~\cite{cha2023learning, mukhoti2023open}, where the number of meta-point is equal to the total number of final pixel embeddings.
A smaller number signifies a reduced number of final semantic units per image,
% and an increased number of relevant patches per set, 
contributing to more consistent segmentation results.
As demonstrated in \cref{tab:meta points}, 
reducing the number of meta-point consistently enhances the segmentation performance.
However, decreasing the number below 36 yields marginal returns in performance, and even leads to a drop in some benchmarks (\eg, Cityscapes and ADE20k). 
Therefore we set the meta-point number as 36 in the aggregation for semantic units. As shown in the penultimate column in \cref{fig:ablation}, the proposed unit achieves more consistent segmentation masks within identical semantic regions, \eg, a more complete mask for the motorbike depicted in the first row.

\subsection{Visualization} 
\begin{figure*}[tb]
\centering
\includegraphics[scale=0.4]{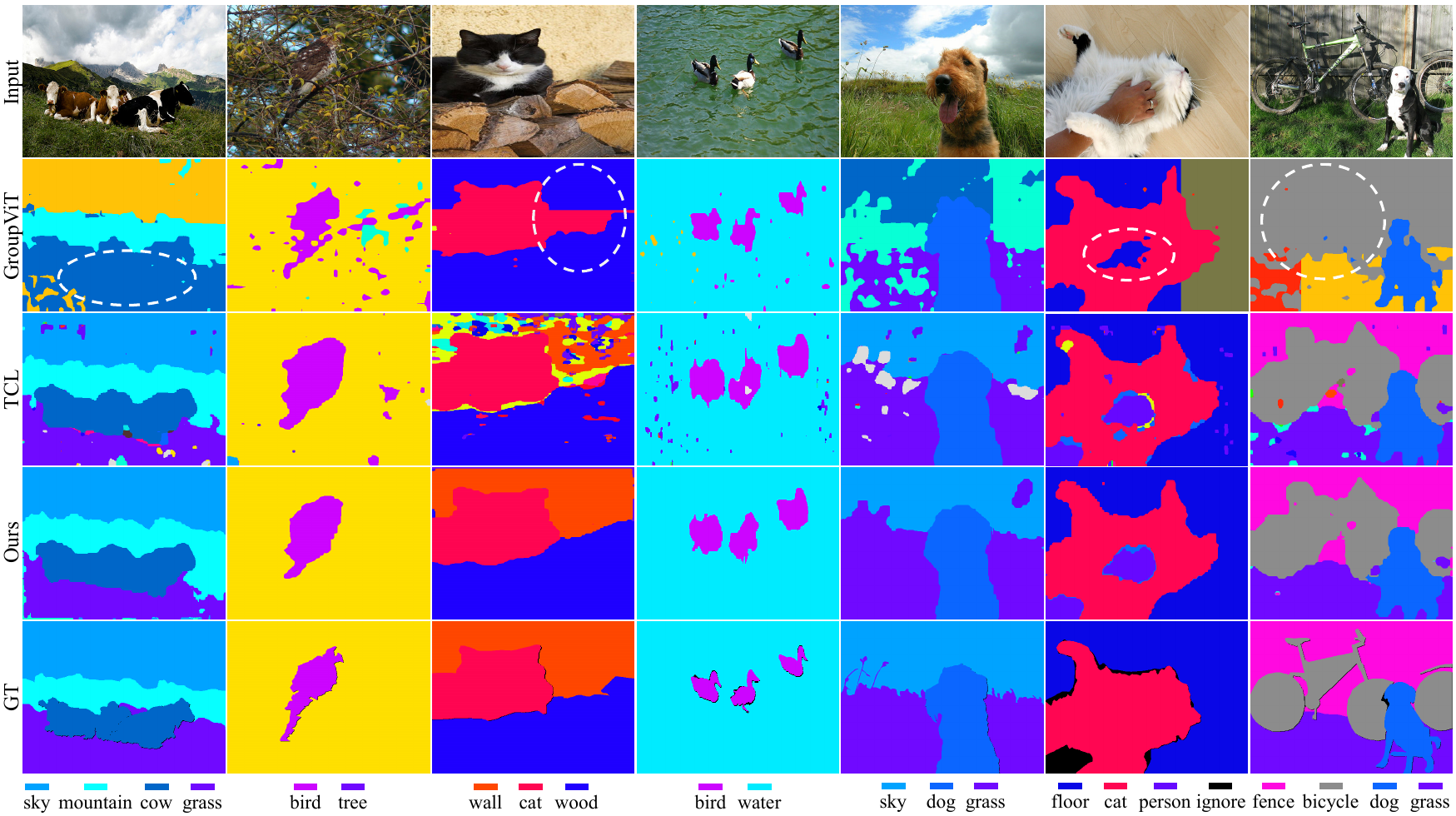}%%%%%%%%%%%%%%%%scale=缩小比例，或者用width=2in
\caption{Qualitative comparison results on PASCAL Context59. Post-processing techniques are not used in all methods. We circle the coarse-clustering and under-segmentation results for GroupViT. Pixel-wise methods like TCL suffer from over-segmentation where a single object is segmented into multiple categories. Our method achieves more accurate and high-quality segmentation masks. The official palette for PASCAL Context59 from MMSegmentation is utilized.}     
\label{fig:result}
% \vspace{-18pt}
\end{figure*}

We qualitatively compare our method with previous state-of-the-art methods in \cref{fig:result}. Specially, we select the group-based GroupViT~\cite{xu2022groupvit} and pixel-wise method TCL~\cite{cha2023learning} to further illustrate the characteristics of different semantic units.
GroupViT encounters inaccurate grouping and under-segmentation problems, while TCL suffers from over-segmentation, particularly with stuff categories. Our method consistently achieves more accurate and high-quality segmentation masks. More qualitative results are provided in Appendix.

\section{Conclusion}
In this paper, we propose an efficient multi-grained cross-modal alignment framework to learn open-vocabulary semantic segmentation without any manual dense annotations. 
Digging into image-text pairs, we innovatively construct multi-grained pseudo semantic correspondence and accomplish object-, region- and pixel-level alignment, explicitly resolving the train-test alignment granularity gap. 
Besides, an adaptive and transferable semantic unit is introduced which cleverly harnesses the advantages of group and pixel units and alleviates their under- and over-segmentation problems in zero-shot predictions. 
With a reduced amount of training data, our method achieves state-of-the-art performance on 8 downstream datasets, highlighting its effectiveness and efficiency.

% \clearpage\mbox{}Page \thepage\ of the manuscript.
% \clearpage\mbox{}Page \thepage\ of the manuscript.
% \clearpage\mbox{}Page \thepage\ of the manuscript.
% \clearpage\mbox{}Page \thepage\ of the manuscript.
% \clearpage\mbox{}Page \thepage\ of the manuscript. This is the last page.
\par\vfill\par

\bibliographystyle{splncs04}
\bibliography{main}

% \clearpage

\end{document}